\documentclass[conference]{IEEEtran}
\IEEEoverridecommandlockouts
\usepackage{cite}
\usepackage{amsmath,amssymb,amsfonts}
\usepackage{algorithmic}
\usepackage{graphicx}
\usepackage{textcomp}
\usepackage{xcolor}
\usepackage{graphics} 
\usepackage{epsfig}
\usepackage{xcolor}
\usepackage{nicefrac}
\usepackage{siunitx}
\usepackage{numprint}
\usepackage{url}
\usepackage{ragged2e}
\def\BibTeX{{\rm B\kern-.05em{\sc i\kern-.025em b}\kern-.08em
    T\kern-.1667em\lower.7ex\hbox{E}\kern-.125emX}}
\newcommand{\osdar}{OSDaR23}
\usepackage{hyperref}
\hypersetup{
    colorlinks=true,
    linkcolor=blue,
    filecolor=blue,      
    urlcolor=blue,
    pdftitle={Overleaf Example},
    pdfpagemode=FullScreen,
    }

\begin{document}

\title{OSDaR23: Open Sensor Data for Rail 2023\\
\thanks{\tiny This work was supported by the DZSF. This is not an official statement, guideline or directive of the German Federal Railway Authority. Human faces and license plates are sufficiently blurred in the images for anonymity. \textcopyright 2023 IEEE. Personal use of this material is permitted. Permission from IEEE must be obtained for all other uses, in any current or future media, including reprinting/republishing this material for advertising or promotional purposes, creating new collective works, for resale or redistribution to servers or lists, or reuse of any copyrighted component of this work in other works. DOI: \href{https://ieeexplore.ieee.org/document/10458449}{10.1109/ICRAE59816.2023.10458449}}
}

\author{
\IEEEauthorblockN{Rustam Tagiew*}
\IEEEauthorblockN{Pavel Klasek}
\IEEEauthorblockN{Roman Tilly}
\IEEEauthorblockA{\textit{German Centre for Rail Traffic Research} \\
\textit{Federal Railway Authority}\\
Dresden, Germany \\
forschung@dzsf.bund.de\\
*Corresponding author}
\and
\IEEEauthorblockN{Martin Köppel}
\IEEEauthorblockN{Patrick Denzler}
\IEEEauthorblockN{Philipp Neumaier}
\IEEEauthorblockN{Tobias Klockau}
\IEEEauthorblockN{Martin Boekhoff}
\IEEEauthorblockA{\textit{Digitale Schiene Deutschland} \\
\textit{DB Netz AG}\\
Berlin, Germany \\
\{martin.koeppel, patrick.denzler,\\
philipp.neumaier\}@deutschebahn.com}
\and
\IEEEauthorblockN{Karsten Schwalbe}
\IEEEauthorblockA{\textit{FusionSystems GmbH}\\
Chemnitz, Germany \\
karsten.schwalbe@fusionsystems.de}
}

\maketitle

\begin{abstract}
To achieve a driverless train operation on mainline railways, actual and potential obstacles for the train's driveway must be detected automatically by appropriate sensor systems. Machine learning algorithms have proven to be powerful tools for this task during the last years. However, these algorithms require large amounts of high-quality annotated data containing railway-specific objects as training data. Unfortunately, all of the publicly available datasets that tackle this requirement are restricted in some way. Therefore, this paper presents \osdar, a multi-sensor dataset of 45 subsequences acquired in Hamburg, Germany, in September 2021, that was created to foster driverless train operation on mainline railways. The sensor setup consists of multiple calibrated and synchronized infrared (IR) and visual (RGB) cameras, lidars, a radar, and position and acceleration sensors mounted on the front of a rail vehicle. In addition to the raw data, the dataset contains \numprint{204091} polyline, polygonal, rectangle, and cuboid annotations in total for 20 different object classes. It is the first publicly available multi-sensor dataset annotated with a variety of object classes that are relevant for the railway context. \osdar, available at \href{data.fid-move.de/dataset/osdar23}{data.fid-move.de/dataset/osdar23}, can also be used for tasks beyond collision prediction, which are listed in this paper.
\end{abstract}

\section{INTRODUCTION}
\label{sec:intro}

In automatic train operation (ATO), technical systems take over tasks that had previously been performed by the operating staff. ATO includes different grades of automation (GoA), up to GoA4 in which the train is fully automated with no staff on board (Tab.~\ref{goas}). 
\begin{figure}[ht!]
  \centering
\includegraphics[width=1\linewidth]{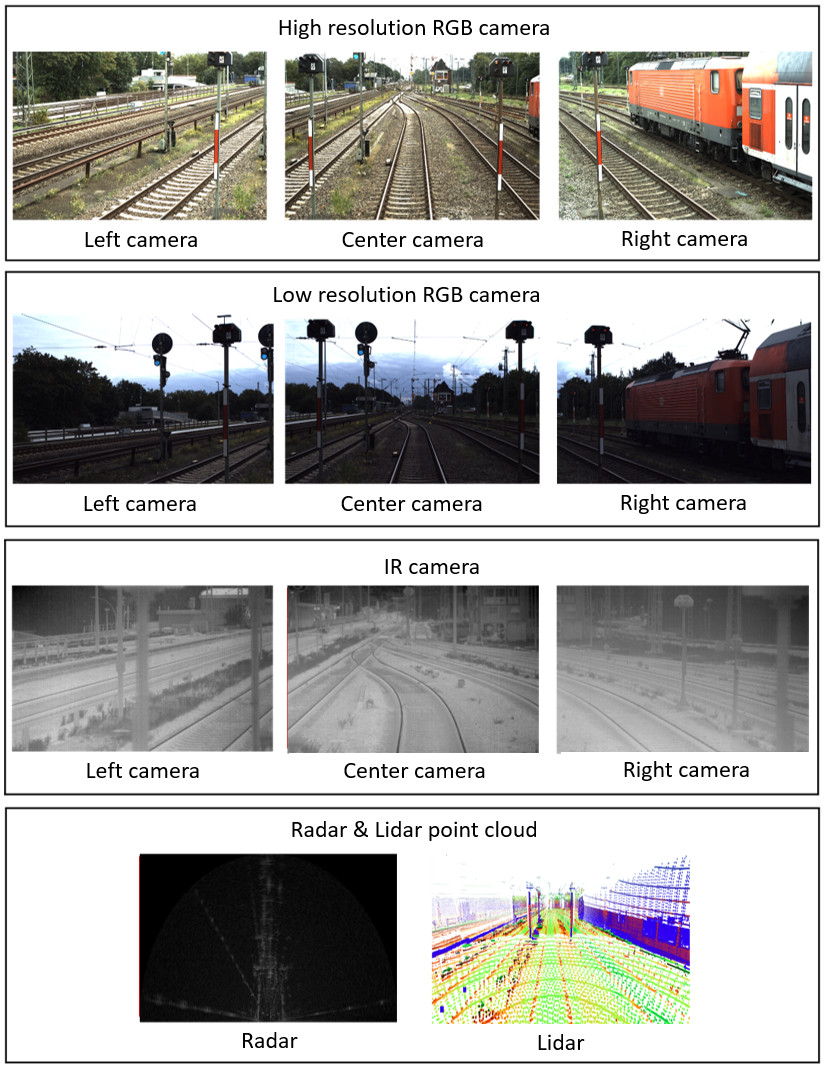}
 \caption{Captured data in a railway environment.}
  \label{overview}
\end{figure}

One of the main tasks of the train driver is to monitor the train's driveway (also known as train's path \cite{ristic2021review}) to predict collisions and act accordingly. This task is rated the most challenging according to a lately conducted rail industry survey \cite{atosensorikreport} and impedes an automation upgrade to GoA3 and GoA4, where no train driver is needed. In this regard, ATO of mainline trains differs from metros. Fully automated metros, such as the Nuremberg U-Bahn, are closed systems meaning that traffic runs in isolated and mostly enclosed environments such as tunnels. Such conditions are also known as ``horizontal lift''. As a result, no object detection is needed and consequently, no computer vision (CV) systems are required on-board \cite{tagiew2021towards}. In contrast, open rail systems are subject to environmental interferences, e.g. at level-crossings or on unfenced tracks. Therefore, various classes of objects may protude in the train's driveway and need to be detected. Hence, replacing human vision with CV systems is a growing area of research for mainline railway.

The usage of machine learning in developing state-of-art CV systems grows \cite{railwayvisonreview}, and therefore large amounts of annotated sensor data. Currently, only a few datasets with objects from the railway environment are available and these datasets are often limited. In order to narrow this gap, this publication presents the ``Open Sensor Data for Rail 2023" (\osdar) \cite{osdaronline,osdar23etr}, a freely accessible annotated multi-sensor dataset containing objects that are relevant to CV systems in the context of the mainline railway and beyond. 
\begin{table}
  \centering
  \caption{Simplified description of GoA from DIN EN 62290-1.}
  \begin{tabular}{|l | c |}
    \hline
    \textbf{GoA}  & \textbf{Train operation}  \\
    \hline
    0 & On-sight with no automation\\ 
    \hline
    1 & Non-automated according to signals\\
      & and with safeguarding in case of overlooked signals      \\ 
    \hline
    2 & Semi-automatic with acceleration and deceleration \\ 
      & controlled by a technical system \\ 
    \hline
    3 & Driverless with a train attendant    \\ 
    \hline
    4 & Unattended without crew on board  \\ 
    \hline
  \end{tabular}
  \label{goas}
\end{table}

While the development of CV-based collision prediction for ATO is the main purpose of \osdar, the dataset may also be used for related tasks and can also serve as a basis for extending open datasets created by others. This dataset can enable developers of CV systems to transfer their algorithms to the railway domain, thus fostering research in this area. 

\osdar~was created in the project ``Aufbereitung von Datensätzen für Anwendungen des automatisierten Fahrens im Eisenbahnbetrieb'' (English: ``Development of Datasets for Applications of Automated Driving in Railway Operations'') conducted by the German Centre for Rail Traffic Research at the Federal Railway Authority (DZSF) using sensor data (cf. Fig.~\ref{overview}) provided by DB Netz AG within the sector initiative of Digitale Schiene Deutschland (DSD). The annotations of this dataset were created by FusionSystems GmbH. The dataset is published together with a research report as well as a labeling guide that specifies how the annotations were created and how future datasets can be annotated similarly. These documents can be obtained from \url{dzsf.bund.de}.

\section{Dataset Requirements for \osdar}
\label{vbcp}

Fully functional CV systems for collision prediction need to include algorithms for the following three sub tasks: obstacle detection, distance estimation, and track detection \cite{ristic2021review}. Given these groups, several requirements for datasets for the development of CV systems can be derived.

First, datasets must contain potential obstacles endogenous to the railway system, such as rail cars and buffer stops, as well as exogenous obstacles, such as pedestrians, road cars, animals, trees, rocks, misplaced drag shoes, fires and many others.

Second, objects are only considered to be obstacles when they are in the train's driveway.
While object detection in the railway context usually refers to objects placed on the tracks, obstacles can belong to rather unusual object types or occur in unusual places as reported cases of bicycles hanging from the catenary illustrate \cite{fahrd5,tagiew2023mainline}.
Therefore, in addition to the existance of objects, their position relative to the driveway needs to be determined. 
The region of interest (ROI) for object detection includes the tracks as well as the 3D tubular space formed by the predicted train's driveway and minimum clearance profile. 
To determine the ROI, the datasets must include tracks and other rail infrastructure objects such as switches and transitions, which allow the prediction of the train's driveway.

Third, compared to road vehicles, braking distances of rail vehicles are about five times larger, and whistling and emergency braking are the only available reactions to obstacles, since evasion is obviously not an option.
Therefore, datasets should include data from sensors that include objects at distances over several hundred meters.

Fourth, object detection performance doesn't only depend on the distance of an object, but also on several other factors, such as the size of the object, the visual contrast of the object compared to the background, the speed of the ego-vehicle, the weather and time of the day, the occlusion of the object by other objects and other visual conditions. 
As a reference, Tab.~\ref{distances} presents median distances for human train driver's performance in detecting objects, which might serve as a benchmark in safety argumentation \cite{tagiew2023mainline}. 
Therefore, another requirement is that datasets need to contain data collected under these different circumstances.

\begin{table}[t]
  \centering
  \caption{Human detection of objects on railway \cite{railwayvisonreview, tagiew2023mainline}.}
  \begin{tabular}{|l|c|}
    \hline
    \textbf{Object} & \textbf{Median distance} \\
    \textbf{(area or size)} & \textbf{of detection in m} \\
    \hline
    $\geq\qty{0.4}{\text{m}^2}$, $\qty{30}{\text{\%}}$ (visual) contrast & $>750$ \\
    $\qty{2}{\text{m}^2}$, $\qty{8}{\text{\%}}$ contrast & $500$  \\
    $\qty{0.4}{\text{m}^2}$, $\qty{8}{\text{\%}}$ contrast & $240$ \\
    $\qty{2}{\text{m}^2}$, $\qty{30}{\text{\%}}$ contrast, at night  & $180$ \\
    $\qty{0.4}{\text{m}^2}$, $\qty{30}{\text{\%}}$ contrast, at night  & $60$ \\
    $\leq\qty{2}{\text{m}^2}$, $\qty{8}{\text{\%}}$ contrast, at night & $<60$ \\
    \cite{polz} & \\
    \hline
    $40\times40\times40$ cm & $250$ \\
    $20\times20\times20$ cm & $175$ \\
    $10\times10\times10$ cm & $50$ \\
    $5\times5\times5$ cm & $<25$  \\
    fluorescent objects at night, $\qty{60}{ \nicefrac{\text{km}}{\text{h}}}$ \cite{itoh} & \\
    \hline
    person in safety jacket & $400$ \\
    passenger car & $300$ \\
    person & $240$ \\
    passenger car at night, person with & $<60$ \\
    and without safety jacket at night & \\
    \cite{mockel2003multi} & \\
    \hline
    tree, $50$-$\qty{70}{\nicefrac{\text{km}}{\text{h}}}$ & $60$ \\
    fallen rock, $20$-$\qty{120}{\nicefrac{\text{km}}{\text{h}}}$ & $30$ \\
    Japanese accident statistics \cite{nakasone2017frontal} & \\ 
    \hline
  \end{tabular}
  \label{distances}
\end{table}

Fifth, depending on GoA and the technical implementation, the task of detecting train signals and their meaning is either conducted by a human, a CV system, or by cab signaling (GoA2). In case of CV systems, the datasets must contain annotated signals. It should also be taken into account that railway signals significantly differ from street signs in shape, colour and placement as well as among railway systems of different countries.

Lastly, it is necessary to predict the future trajectory of objects in order to determine whether a collision is likely to happen or not. This requires tracking objects over time. Therefore, the datasets need to contain tracking identifiers that allow mapping different annotations to the same real-world object.

\section{Existing Datasets for CV in Railway}
\label{rvd}
\begin{table}
  \centering
  \caption{Existing Public Single-Sensor Datasets for CV in Railway}
  \begin{tabular}{|l|r|r|r|}
    \hline
    \textbf{Dataset}& \textbf{Year} & \textbf{Frames} & \textbf{RGB image format} \\
    \hline
     RailSem19\cite{railsem19}& 2019 & $\numprint{8500}$   & variable \\
     FRSign\cite{FRSign}      & 2020 & $\numprint{105352}$ & $2048\times 1536$, $1920\times 1200$\\
     RAWPED\cite{rawped}      & 2020 & $\numprint{26000}$  & variable \\  
     Rail-DB\cite{raildb}     & 2022 & $\numprint{7432}$   & $800\times 288$ \\
     RailSet\cite{railset}    & 2022 & $\numprint{6600}$   & variable \\
     GERALD\cite{gerald}      & 2023 & $\numprint{5000}$   & $1280\times 720$, $1920\times 1080$\\
    \hline
  \end{tabular}
  \label{exdrv}
\end{table}
To the best of the authors' knowledge, datasets listed in Tab.~\ref{exdrv} are the only CV datasets recorded by frontal on-board sensors of a railway train including object annotations that have explicitly been published for free use by the research community. They contain annotated single RGB-camera frames from video sequences. RailSem19 contains frames of railway and tram scenes from $38$ countries. It contains annotations in form of geometric shapes and dense pixel-wise semantic segmentation for trains, switches, switch states, platforms, buffer stops, rail traffic signs and railway signals. RailSet is similar in its annotation to RailSem19 and contains annotated railway scenes, both under regular and defective conditions, such as rails discontinuity and hole anomalies. Rail-DB contains annotated recordings of high-speed rail infrastructure.

FRSign is a dataset of frames annotated with boxes of French railway signals in different signal states and GERALD a dataset of frames containing German signals. RAWPED contains frames with box annotations for pedestrians. In addition to open datasets, there are also datasets that have been used in research but have not (yet) been published, such as, the RAILOD dataset of $\numprint{4651}$ manually annotated single RGB-camera frames from $6$ scenes \cite{guan2022lightweight}.

\section{Motivation for \osdar}
\label{propdataset}
\osdar~\cite{osdaronline} is a manually annotated open dataset with a multi-sensor setup for the railway environment. The multi-sensor setup includes cameras, lidars, a radar as well as position and acceleration sensors. In contrast, available open datasets for railway focus mainly on camera images. The goal of creating \osdar~was to advance the development of AI algorithms for the higher GoA of mainline railways, in terms of object detection. Another related goal is making developers of AI systems more familiar with the railway context and to support standardization activities for the safety approval process \cite{ATOdataset}.
\section{Multi-Sensor System}
\label{fmss}
\begin{figure}[t]
  \centering
\includegraphics[width=1\linewidth]{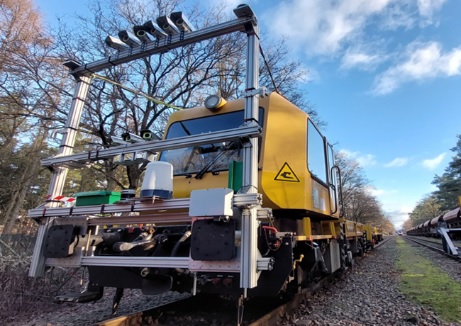}
 \caption{The utilized vehicle, with the mounted sensor setup \cite{osdar23etr}. 
  }
  \label{gaf}
\end{figure}
\osdar~was recorded on several data collection runs in September 2021 in Hamburg, Germany, by DB Netz AG. 
The data was then selected in a way that it covers many different situations in regular train operation. Additionally, some special situations and objects, e.g. flames and smoke, were staged.

A DZSF research project on sensor technology for driverless trains has shown that multi-sensor systems can guarantee a complete environmental perception \cite{atosensorikreport}. Although the optimal sensor setup is still a subject of research, lidar, radar, RGB and IR cameras have proven to be particularly useful. Due to the long range requirements resulting from long stopping distances, cameras with high resolutions and narrow fields of view (FoV) should be used. Narrow FoVs and the presence of curved tracks require multiple cameras positioned at different angles relative to the vehicle.

\begin{figure}[t]
  \centering
  \includegraphics[width=1\linewidth]{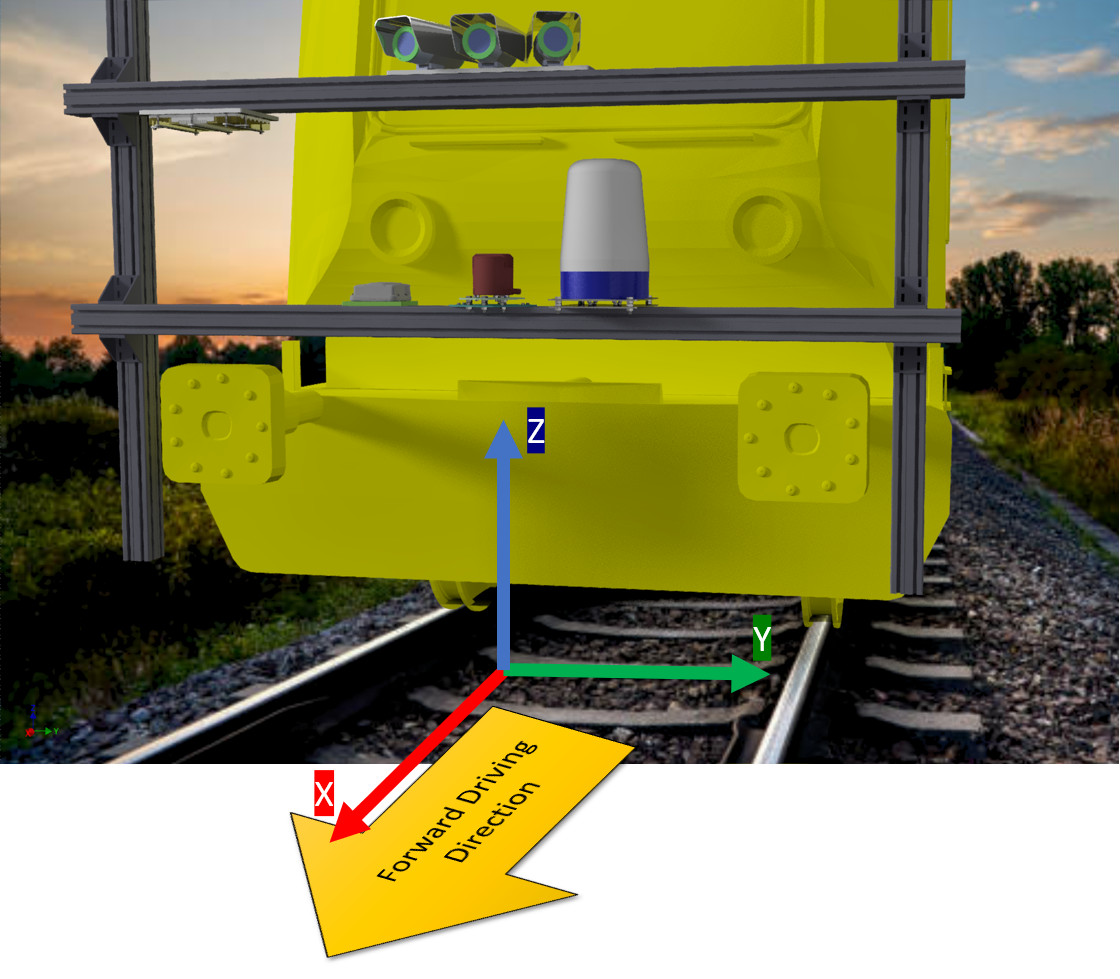}
  \caption{The utilized reference coordinate system.}
  \label{coordSyst}
\end{figure}

The vehicle utilized for data collection was a track working vehicle with attached profiles for the multi-sensor system. Fig.~\ref{gaf} shows the sensors at the front of the vehicle mounted on the profiles. The sensor configuration included a total of six RGB cameras, three IR cameras, six lidar sensors with different ranges and field of view (FoV) coverage, a 2D radar sensor, and position and acceleration sensors (GNSS+IMU). The camera, lidar, and radar data are shown in Fig.~\ref{overview}. The cameras are divided into multiple sets of $3$ cameras each, aligned like a trident, diagonally left, center and diagonally right in the direction of travel. The main technical specifications of the sensors are listed in Tab.~\ref{msystem}. The sensors were calibrated according to the reference coordinate system shown in Fig.~\ref{coordSyst}. The FoV areas of the sensors covered the area in front and partially to the side of the vehicle and overlap each other. 

The sensors were synchronized at a frame rate of $\qty{10}{\text{Hz}}$ based on the acquisition time of each sensor. The sensors used the Precision Time Protocol (PTP) in order to synchronize their clocks. As the radar has a capturing rate of $\qty{4}{\text{Hz}}$ only, some radar images are duplicated.
The union of frames from all sensors at a given point in time forms a multi-sensor frame (m-frame). For the dataset, the point clouds of the six lidars were ego-motion compensated and merged into a single point cloud per m-frame. 

\begin{table}[t]
  \centering
  \caption{Multi-sensor system}
  \begin{tabular}{|l|c|}
\hline
\multicolumn{2}{|l|}{\textbf{Three 12MP RGB cameras}} \\
\hline
Type & Teledyne GenieNano 5GigE C4040 \\
Sensor data & RGB images (8 Bit, PNG) \\
Resolution & $\numprint{4112}\times\numprint{2504}$ px \\
Sampling frequency & $\qty{10}{\text{Hz}}$ (synchronized) \\
Alignment & trident \\
\hline
\multicolumn{2}{|l|}{\textbf{Three 5MP RGB cameras}} \\
\hline
Type & Teledyne GenieNano C2420\\
Sensor data & RGB images (8 Bit, PNG) \\
Resolution & $\numprint{2464}\times\numprint{1600}$ px \\
Sampling frequency & $\qty{10}{\text{Hz}}$ (synchronized) \\
Alignment & trident \\
\hline        
\multicolumn{2}{|l|}{\textbf{Three IR cameras}} \\
\hline
Type & Teledyne Calibir DXM640 \\
Sensor data & grayscale images (8 Bit, PNG) \\
Resolution & $640\times 480$ px \\
Sampling frequency & $\qty{10}{\text{Hz}}$ (synchronized) \\
Alignment & trident \\
\hline  
\multicolumn{2}{|l|}{\textbf{Three long-range lidars}} \\
\hline
Type & Livox Tele-15 \\
Sensor data & 3D point cloud (PCD) \\
Total sampling points & $\numprint{50000}$ - $\numprint{84000}$ points per frame \\
Sampling frequency & $\qty{10}{\text{Hz}}$ (synchronized) \\
\hline           
\multicolumn{2}{|l|}{\textbf{One medium-range lidar}} \\
\hline
Type & HesaiTech Pandar64 \\
Sensor data & 3D point cloud (PCD) \\
Total sampling points & $\numprint{60000}$ - $\numprint{115200}$ points per frame \\
Sampling frequency & $\qty{10}{\text{Hz}}$ (synchronized) \\
\hline           
\multicolumn{2}{|l|}{\textbf{Two short-range lidars}} \\
\hline
Type & Waymo Honeycomb \\
Sensor data & 3D point cloud (PCD) \\
Total sampling points & $\numprint{20000}$ - $\numprint{40000}$ points per frame \\
Sampling frequency & $\qty{10}{\text{Hz}}$ (synchronized) \\
\hline           
\multicolumn{2}{|l|}{\textbf{One radar}} \\
\hline
Type & Navtech CIR204/H \\
Sensor data & grayscale images (8 bit, PNG), \\
            &cartesian bird's eye view \\
Resolution & $\numprint{2856}\times\numprint{1428}$ px \\
Sampling frequency & 4 Hz (synchronized) \\
\hline        
\multicolumn{2}{|l|}{\textbf{Global navigation satellite system (GNSS) sensor with}} \\
\multicolumn{2}{|l|}{\textbf{inertial measurement unit (IMU)}} \\
\hline
Type & NovAtel PwrPAk7D-E1\\
Sensor data GNSS & latitude and longitude in WGS84 \\
Sensor data IMU & linear and rotatory acceleration\\
Sampling frequency & $100$/$\qty{10}{\text{Hz}}$ \\
\hline
  \end{tabular}
  \label{msystem}
\end{table}

\section{Annotation Specification}
\label{acf}
In addition to the raw sensor data, the annotations are the core part of OSDaR23. A labeling guide forms the basis for the manual creation of annotations by the annotators. This guide refers to the described sensor data from IR/RGB cameras, the radar and point cloud data from the lidars.

The choice of relevant objects for annotation is motivated by the tasks of a contactless perception system replacing a human driver as described in Sec.~\ref{vbcp}, whereby no specific operational design domain is chosen. The task of obstacle detection requires the annotation of endangered objects such as people and animals as well as endangering objects such as road and rail vehicles, drag shoes etc. The ROI detection task requires the annotation of infrastructure. Signal detection requires the annotation of signals.

In the camera and radar frames, axis-parallel and rotated rectangles, polylines and polygons are used as two-dimensional (2D) annotation geometries. The rectangles and polygons enclose the annotated objects in the sensor images as accurately as possible. The polylines are used to annotate tracks and transitions -- they follow the outer rail edges. 
In the lidar point clouds, three-dimensional (3D) cuboids and polylines are used, as well as semantic segmentations. Cuboids enclose objects and polylines mark contours analogous to the 2D annotations. In semantic segmentations, individual lidar points are assigned to an object.

Using these geometries, objects of $20$ different classes from four different categories were annotated. The classes for dynamic objects are person, crowd, train, wagons, bicycle, group of bicycles, motorcycle, road vehicle, animal, group of animals, wheelchair, and drag shoe. The railway related objects are track, switch, and transition. Static objects are catenary pole, signal pole, signal, signal bridge, and buffer stop. Special classes are flame and smoke. As the sensor data were recorded in Germany, only German railway signals are part of the dataset.
All object annotations have tracking IDs by which they can be assigned to the same physical object in the real world in all m-frames of a sequence. 

The annotations are provided in JSON-files that follow the RailLabel JSON schema. The RailLabel schema is a sub-schema of the universally usable ASAM OpenLABEL standard developed by the Association for Standardization of Automation and Measuring systems \cite{asam}.
The RailLabel scheme specifies options that were kept open in the standard ASAM OpenLABEL format. 
The development software kit for this dataset is online available on \url{github.com/DSD-DBS/raillabel}. 

\begin{figure*}
    \includegraphics[width=.46\textwidth]{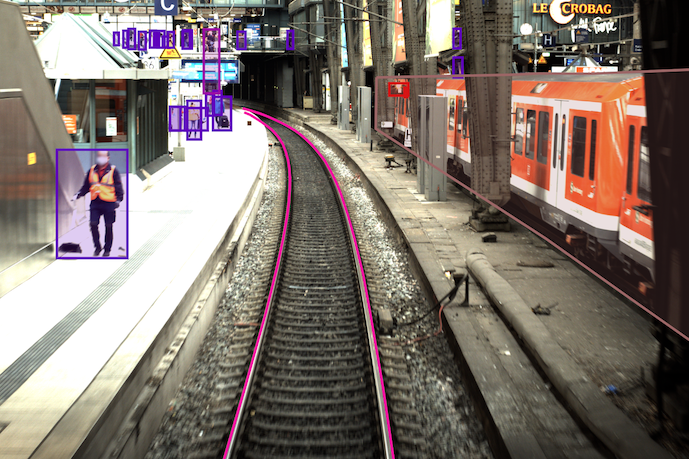}\hfill
    \includegraphics[width=.46\textwidth]{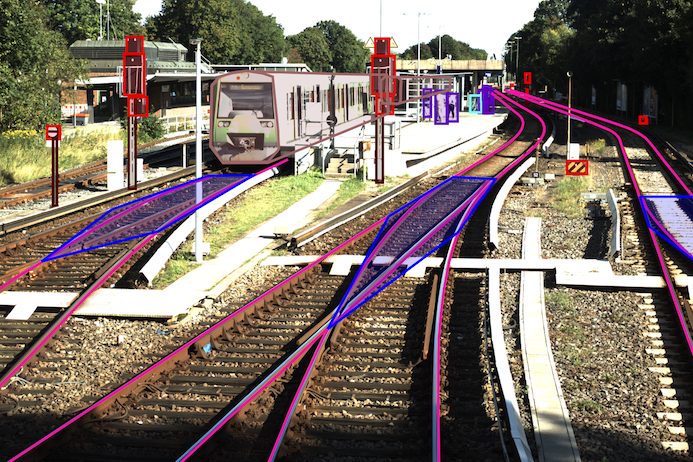}\hfill
    \\[\smallskipamount]
    \includegraphics[width=.46\textwidth]{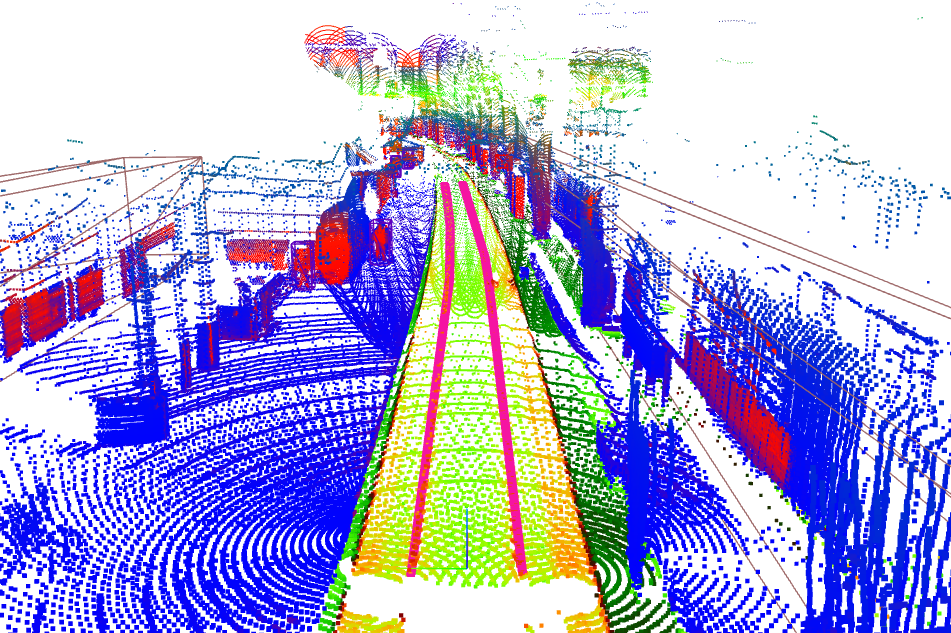}\hfill
    \includegraphics[width=.46\textwidth,trim={0 0 0 0.3cm},clip]{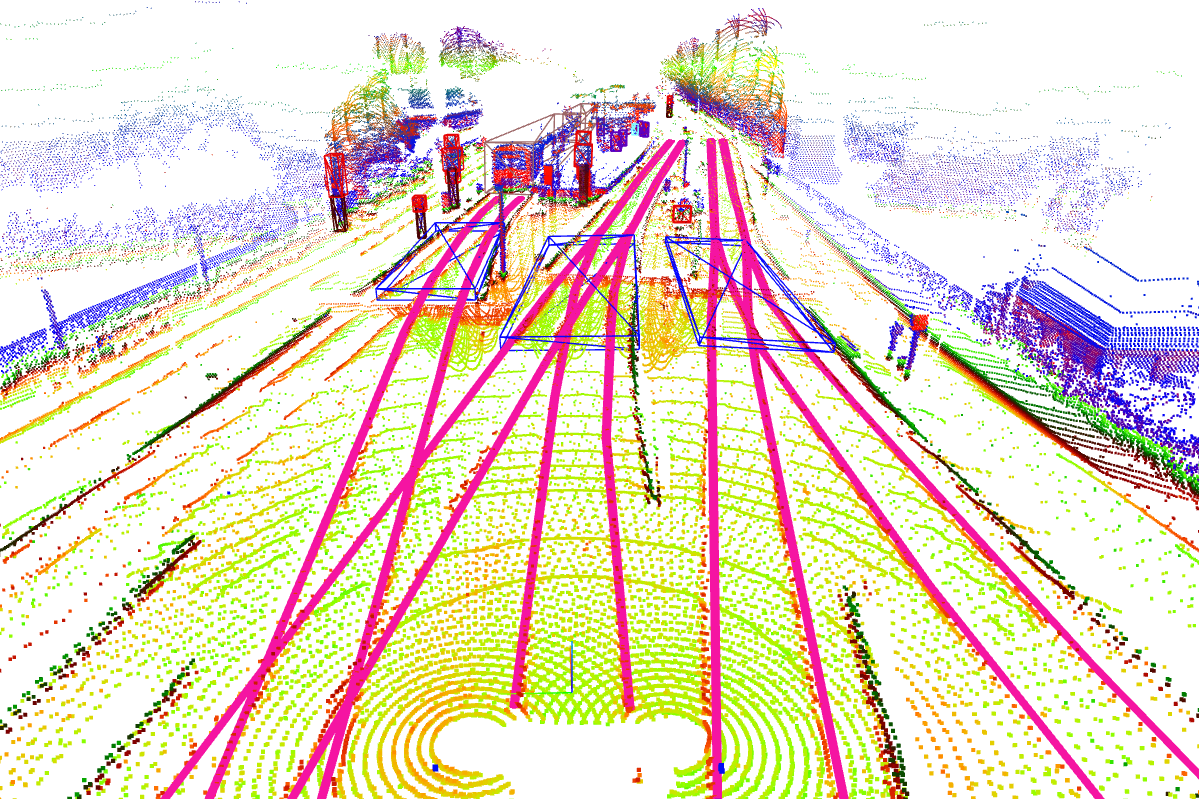}\hfill
    \\[\smallskipamount]
    \includegraphics[width=.46\textwidth,trim={0 0 0 0.1cm},clip]{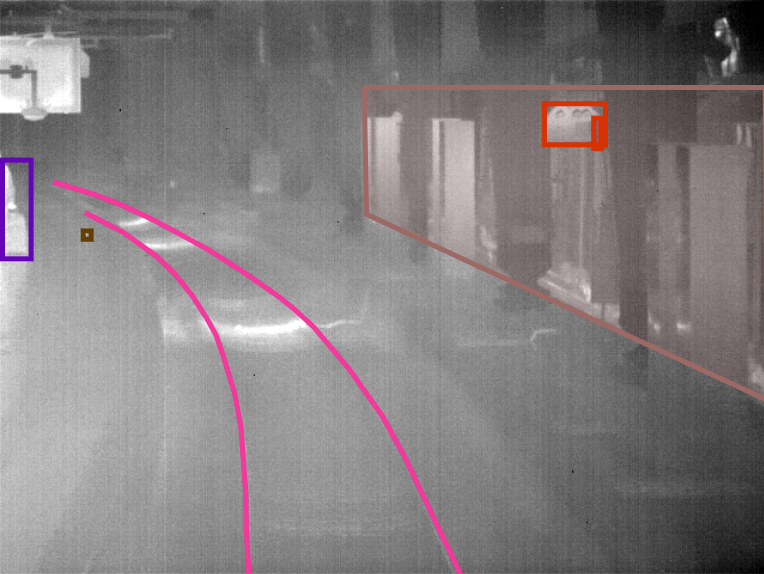}\hfill
    \includegraphics[width=.46\textwidth]{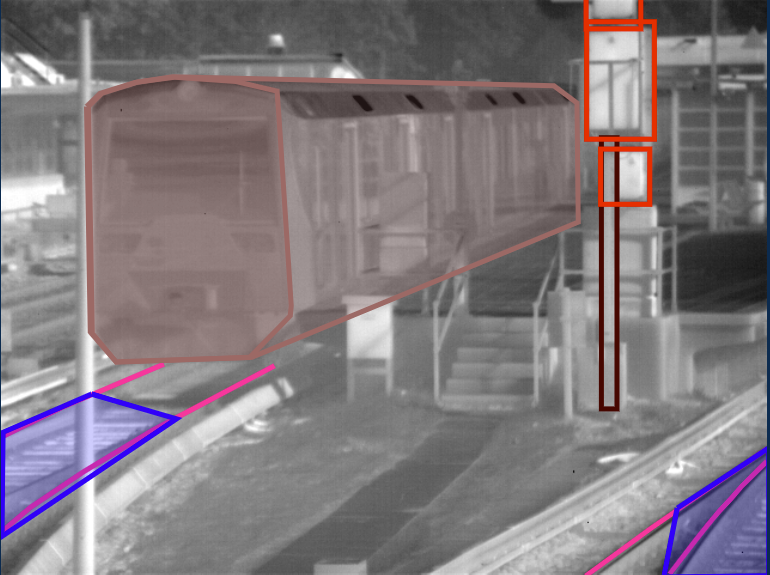}\hfill
    \\[\smallskipamount]
    \includegraphics[width=.46\textwidth]     {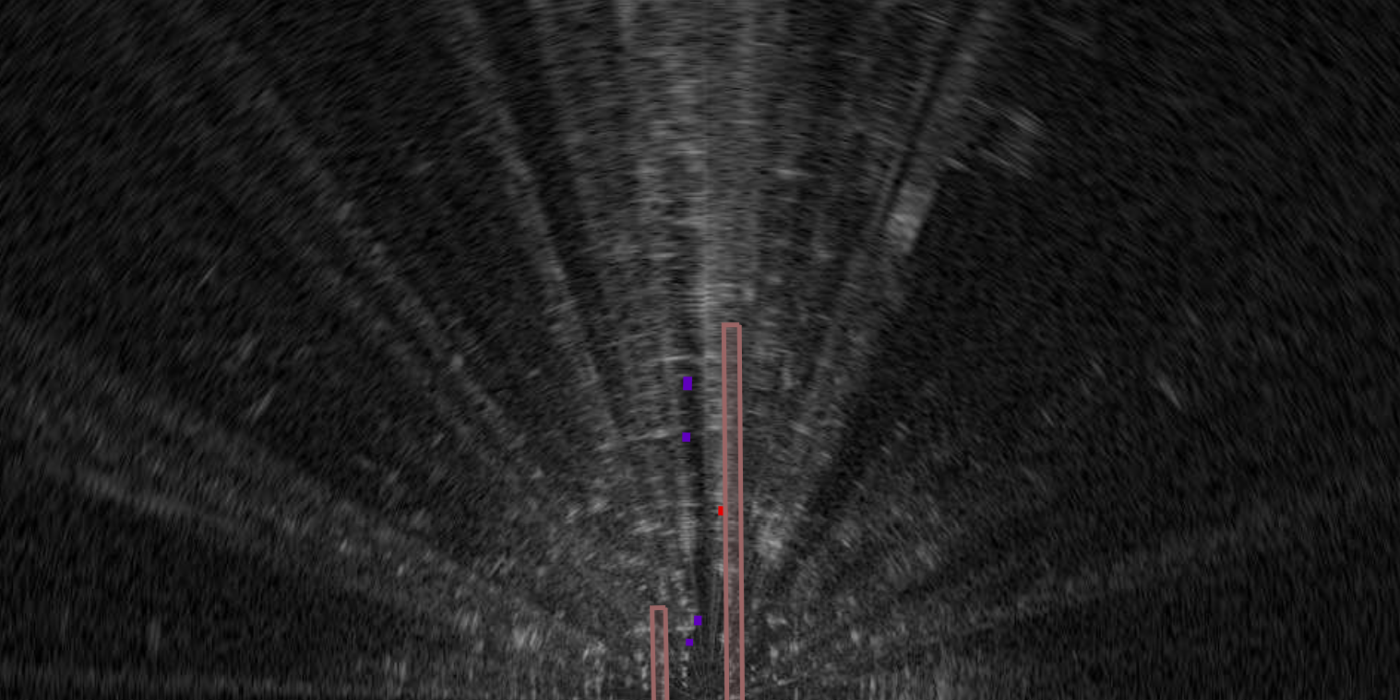}\hfill
    \includegraphics[width=.46\textwidth]{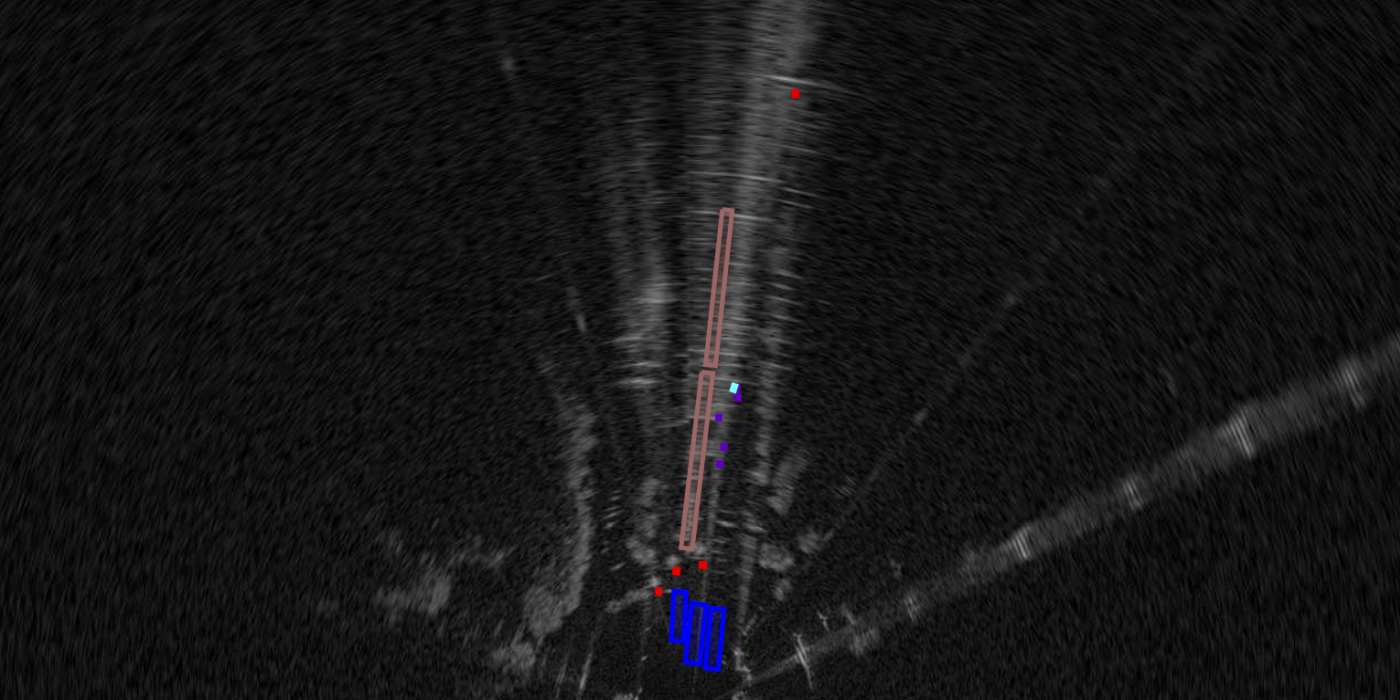}\hfill
    \caption{Examples for annotated sensor data from \osdar. Left column: Sequence inside Hamburg Main Station. Right column: Sequence at Station Ohlsdorf. 
    First Row: RGB center camera. Second Row: Merged lidar point cloud. Third row: IR center camera. Last Row: Radar (zoomed). In the sensor frames different annotations are visualized: Rectangles and cuboids for persons (dark blue), switches (blue), signals (red), signal poles (dark brown), polylines for rails (pink) and polygons for trains beige).}\label{fig:SensorExamples}
\end{figure*}

Details on class attributes, annotation rules and the storage format are presented in the project report and in the separately published labeling guide. Fig.~\ref{fig:SensorExamples} shows two examples for the sensor data and its associated annotations. In the left column a recording of the main station in Hamburg is presented. In the right column a recording in the proximity to station Hamburg-Ohlsdorf is shown. The first row shows the images from the center camera. In the second row the ego-motion compensated and merged point clouds of the lidars are shown. The third row presents the data of the central IR camera. In the last row the outcome of the radar is pictured. Please note, that the radar images are zoomed in due to the long viewing range of the radar. 

\section{\osdar~Statistics}
\label{dss}
The annotated dataset comprises $21$ sequences, divided into $45$ subsequences, with a total of $\numprint{1534}$ annotated m-frames and $\numprint{204091}$ annotation objects. Since an m-frame is composed of nine camera frames, one radar frame and one lidar frame, the total number of individual sensor frames is $\numprint{1534}\cdot11 = \numprint{16874}$. An annotation object refers to the annotation of a physical real-world object captured in a sensor at a point in time. A real-world object is typically captured by multiple sensors and over multiple points in time and is therefore represented in multiple annotation objects. Additionally, for each annotation, the values for several class-specific attributes are given.

To cover a wide variety of object classes and environments while enabling the development of object tracking, the dataset contains several shorter sequences of different locations and situations with ten m-frames each as well as some longer sequences of $40$ to $100$ m-frames. The most common annotation objects are persons such as passengers or staff. These are followed by static objects such as signals, catenary poles, tracks, signal poles, buffer stops, and dynamic objects such as road vehicles and trains. Tab.~\ref{annpo} gives an overview of the distribution of the annotation objects.

\osdar~can be divided into training, test and validation sets. Our suggested division is set in a way, that $\sim 70\%$ of the m-frames can be used for training and $\sim 15\%$ for test and validation. In the following the division is outlined: 

\raggedright{
\textbf{Training dataset}: 
1\_calibration\_1.2, 3\_fire\_site\_3.1, 3\_fire\_site\_3.3, 4\_station\_pedestrian\_bridge\_4.3, 5\_station\_bergedorf\_5.1, 6\_station\_klein\_flottbek\_6.2, 8\_station\_altona\_8.1, 8\_station\_altona\_8.2, 9\_station\_ruebenkamp\_9.1, 12\_vegetation\_steady\_12.1, 14\_signals\_station\_14.1, 15\_construction\_vehicle\_15.1, 20\_vegetation\_squirrel\_20.1, 21\_station\_wedel\_21.1, 21\_station\_wedel\_21.2, \\
\textbf{Test dataset}:
1\_calibration\_1.1, 3\_fire\_site\_3.2, 4\_station\_pedestrian\_bridge\_4.1, 4\_station\_pedestrian\_bridge\_4.4,
5\_station\_bergedorf\_5.2, 7\_approach\_underground\_station\_7.1,
7\_approach\_underground\_station\_7.3, 8\_station\_altona\_8.3, 9\_station\_ruebenkamp\_9.2, 9\_station\_ruebenkamp\_9.6,
10\_station\_suelldorf\_10.1, 13\_station\_ohlsdorf\_13.1,
16\_under\_bridge\_16.1, 17\_signal\_bridge\_17.1,
19\_vegetation\_curve\_19.1 \\
\textbf{Validation dataset}:
2\_station\_berliner\_tor\_2.1,
3\_fire\_site\_3.4,
4\_station\_pedestrian\_bridge\_4.2,
4\_station\_pedestrian\_bridge\_4.5
6\_station\_klein\_flottbek\_6.1
7\_approach\_underground\_station\_7.2
9\_station\_ruebenkamp\_9.3,
9\_station\_ruebenkamp\_9.4,
9\_station\_ruebenkamp\_9.5,
9\_station\_ruebenkamp\_9.7,
11\_main\_station\_11.1,
14\_signals\_station\_14.2,
14\_signals\_station\_14.3,
18\_vegetation\_switch\_18.1,
21\_station\_wedel\_21.3 \\
}
\setlength\parindent{1em}
\justifying
\begin{table}[t]
\centering
\caption{Number of annotations per object class}
\begin{tabular}{|l|r|l|r|}
\hline  
\textbf{Object Class} & \textbf{Count} & \textbf{Object Class} & \textbf{Count} \\
\hline
person & $\numprint{73421}$ & bicycle & $\numprint{1779}$ \\
signal & $\numprint{32790}$ & crowd & $\numprint{1352}$\\
catenary pole & $\numprint{27706}$ & bicycles & $644$ \\
track & $\numprint{18543}$ & transition & $636$ \\
signal pole & $\numprint{14374}$ & flame & $410$\\
road vehicle & $\numprint{12669}$ & signal bridge & $312$\\
train & $\numprint{8290}$ & smoke & $188$\\
buffer stop & $\numprint{4539}$ & wagons & $110$\\
animal & $\numprint{3288}$ & drag shoe & $79$\\
switch & $\numprint{2947}$ & motorcycle & $14$ \\
\hline
\end{tabular}
\label{annpo}
\end{table}

\section{Limitations}
\label{Limitations}

In the following, the limitations of \osdar~are the explained.

First, the dataset does not include an associated digital map. Nonetheless, a digital map can be created from the annotated data using visual odometry or self-localization.

Second, while signals are annotated, they are not associated with the track to which they belong. Since exceptions to the rules for signal placement exist, these associations can not be easily derived \cite{petrovic2022integration}. Further, \osdar~does not provide the classification of signal states. 

Third, automation of shunting operation in shunting yards requires recordings and annotations of special signals such as fouling point indicators at the railway switches, which \osdar~does not provide.  

Lastly, the visual inspection of the infrastructure and vehicle is an important task to prevent accidents additionally to predictive maintenance. Damaged infrastructure and slipping load of crossing trains must be detected to induce emergency braking. However, such cases are not covered by \osdar.

\section{Conclusion and Future Work}
\label{futurework}
The open multi-sensor dataset \osdar~\cite{osdaronline} for driverless operation in the railway environment is presented, probably the first of its kind. Multiple calibrated synchronized IR/RGB cameras, lidars, a radar and GNSS+IMU sensors are mounted on the front of a track work vehicle during data acquisition. The dataset contains $\numprint{204091}$ annotations for $20$ different object classes. Developers can now use an industrial-grade dataset to create AI models for object detection, ROI determination, and distance estimation required for collision prediction by state-of-the-art perception systems. The dataset can be extended with user-specific complementary data like objects intruding into the railways.

Today, there are more than $200$ known methods for pedestrian motion prediction \cite{rudenko2020human}, which can be tested on \osdar. Detecting pedestrians is a very important task as trespassing railways is a life-threatening criminal offense that also obstructs train operation.

Further datasets beyond \osdar~will be required to develop CV systems fulfilling high safety requirements of ATO. They should include different sensor configurations, a quantitative increase of data, qualitative expansion of the object classes, their attributes, geometries, environments, and situations as well as recordings of unusual, (anticipated) critical and incident events. The DZSF continues the activities in processing and publishing annotated sensor data with the support of DB Netz AG. 

\osdar~will be integrated in the DSD Data-Factory \cite{datafactory}. The Data-Factory is a platform for the systematic provision and processing of sensor data for the development of AI functions and for the simulation of photorealistic scenarios including reasonable trajectories of the relevant objects as well as artificial sensor data. 

\osdar~and its annotation specification can serve as a reference as well as a basis for extensions, which will fill the gaps described in Sec.~\ref{Limitations}. Developers of CV systems that work in related fields of research like visual infrastructure inspection \cite{liu2019review}, automated security surveillance \cite{zaman2019artificial}, train door operation, digital railway mapping etc. performed on-board and off-board might take advantage of these extensions and contribute to the development of further datasets. Similarly, developers of multi-sensor data generation systems for railways \cite{damico2023trainsim} might also improve the performance of their system by using \osdar. All stakeholders in the rail sector and beyond are invited to participate in this effort and, if possible, publicate new datasets to achieve a broad research and development community.

\section*{ACKNOWLEDGMENT}
The authors thank their colleagues Dr. Kai Hofmann (DZSF), Florian Reiniger, Markus-Franz Ziegler, Volker Eiselein, Arne Jacobs (all DB Netz AG) and Michael Scheithauer, Kai-Uwe Kaden and Kevin Förster (all FusionSystems GmbH) for their support in the project, as well as the Vicomtech Research Foundation (\url{vicomtech.org}) for providing the \mbox{WebLabel} Player which can be used to visualize the dataset.

\bibliographystyle{IEEEtran}
\bibliography{osdar23}
\end{document}